\documentclass[letterpaper,twocolumn,10pt]{article}
\usepackage{usenix2019_v3}

\usepackage{tikz}
\usepackage{amsmath}

\usepackage{filecontents}

\begin{document}

\date{}

\title{Deep Learning in the Era of Edge Computing: Challenges and Opportunities}

\author{
{\rm Mi Zhang\textsuperscript{$1$}, Faen Zhang\textsuperscript{$2$}, Nicholas D. Lane\textsuperscript{$3$}, Yuanchao Shu\textsuperscript{$4$}, Xiao Zeng\textsuperscript{$1$}, Biyi Fang\textsuperscript{$1$}, Shen Yan\textsuperscript{$1$}, Hui Xu\textsuperscript{$2$}} \\
\textsuperscript{$1$}Michigan State University, \textsuperscript{$2$}AInnovation, \textsuperscript{$3$}Oxford University, \textsuperscript{$4$}Microsoft Research 
\\
mizhang@msu.edu, zhangfaen@ainnovationn.com, nicholas.lane@cs.ox.ac.uk, yuanchao.shu@microsoft.com, 
\\ \{zengxia6, fangbiyi, yanshen6\}@msu.edu, xuhui@ainnovationn.com
} 

\maketitle

\begin{abstract}
The era of edge computing has arrived.
Although the Internet is the backbone of edge computing, its true value lies at the intersection of gathering data from sensors and extracting meaningful information from the sensor data.
We envision that in the near future, majority of edge devices will be equipped with machine intelligence powered by deep learning. 
However, deep learning-based approaches require a large volume of high-quality data to train and are very expensive in terms of computation, memory, and power consumption.
In this chapter, we describe eight research challenges and promising opportunities at the intersection of computer systems, networking, and machine learning. 
Solving those challenges will enable resource-limited edge devices to leverage the amazing capability of deep learning.
We hope this chapter could inspire new research that will eventually lead to the realization of the vision of intelligent edge.

%
\end{abstract}

\section{Introduction}
\label{sec.intro}


Of all the technology trends that are taking place right now, perhaps the biggest one is \textit{edge computing}~\cite{shi2016edge,shi2016promise}. 
It is the one that is going to bring the most disruption and the most opportunity over the next decade.
Broadly speaking, edge computing is a new computing paradigm which aims to leverage devices that are deployed at the Internet's edge to collect information from individuals and the physical world as well as to process those collected information in a distributed manner~\cite{satyanarayanan2017emergence}.
These devices, referred to as \textit{edge devices}, are physical devices equipped with sensing, computing, and communication capabilities. 
Today, we are already surrounded by a variety of such edge devices:
our mobile phones and wearables are edge devices; home intelligence devices such as Google Nest and Amazon Echo are edge devices; autonomous systems such as drones, self-driving vehicles, and robots that vacuum the carpet are also edge devices.
These edge devices continuously collect a variety of data including images, videos, audios, texts, user logs, and many others with the ultimate goal to provide a wide range of services to improve the quality of people's everyday lives.

Although the Internet is the backbone of edge computing, the true value of edge computing lies at the intersection of gathering data from sensors and extracting meaningful information from the collected sensor data. 
Over the past few years, deep learning (i.e., Deep Neural Networks (DNNs))~\cite{lecun2015deep} has become the dominant data analytics approach due to its capability to achieve impressively high accuracies on a variety of important computing tasks such as speech recognition \cite{hinton2012deep}, machine translation \cite{bahdanau2014neural}, object recognition \cite{krizhevsky2012imagenet}, face detection \cite{taigman2014deepface}, sign language translation \cite{fang2017deepasl}, and scene understanding \cite{zhou2014learning}.
%
Driven by deep learning's splendid capability, companies such as Google, Facebook, Microsoft, and Amazon are embracing this technological breakthrough and using deep learning as the core technique to power many of their services.

Deep learning models are known to be expensive in terms of computation, memory, and power consumption~\cite{he2016deep, simonyan2014very}.
As such, given the resource constraints of edge devices, the \textit{status quo} approach is based on the cloud computing paradigm in which the collected sensor data are directly uploaded to the cloud; and the data processing tasks are performed on the cloud servers where abundant computing and storage resources are available to execute the deep learning models.
%
Unfortunately, cloud computing suffers from three key drawbacks that make it less favorable to applications and services enabled by edge devices.
First, data transmission to the cloud becomes impossible if the Internet connection is unstable or even lost. 
Second, data collected at edge devices may contain very sensitive and private information about individuals. Directly uploading those raw data onto the cloud constitutes a great danger to individuals' privacy.
%
Most importantly, as the number of edge devices continues to grow exponentially, the bandwidth of the Internet becomes the bottleneck of cloud computing, making it no longer be feasible or cost effective to transmit the gigantic amount of data collected by those devices to the cloud.


In this book chapter, we aim to provide our insights for answering the following question: \textit{can edge computing leverage the amazing capability of deep learning?}
As computing resources in edge devices become increasingly powerful, especially with the emergence of Artificial Intelligence (AI) chipsets, we envision that in the near future, majority of the edge devices will be equipped with machine intelligence powered by deep learning.
The realization of this vision requires considerable innovation at the intersection of computer systems, networking, and machine learning.
In the following, we describe eight research challenges followed by opportunities that have high promise to address those challenges.
We hope this book chapter act as an enabler of inspiring new research that will eventually lead to the realization of the envisioned intelligent edge.

\section{Challenges and Opportunities}
\label{sec.challenge}


\vspace{1mm}
\subsection{\textbf{Memory and Computational Expensiveness of DNN Models}}

DNN models that achieve state-of-the-art performance are memory and computational expensive.
To illustrate this, Table \ref{tab.CNN_examples} lists the details of some of the most commonly used DNN models.
As shown, these models normally contain millions of model parameters and consume billions of floating-point operations (FLOPs).
This is because these DNN models are designed for achieving high accuracy without taking resources consumption into consideration.
%
Although computing resources in edge devices are expected to become increasingly powerful, their resources are way more constrained than cloud servers. 
Therefore, filling the gap between high computational demand of DNN models and the limited computing resources of edge devices represents a significant challenge.

\begin{table}[b]
	\centering
		\scalebox{0.75}{
	\begin{tabular}{|c|c|c|c|c|c|}
		\hline
		\textbf{DNN} & \textbf{\begin{tabular}[c]{@{}c@{}}Top-5 Error\\ (\%)\end{tabular}} & \textbf{\begin{tabular}[c]{@{}c@{}}Latency\\ (ms)\end{tabular}} & \textbf{Layers} & \textbf{\begin{tabular}[c]{@{}c@{}}FLOPs\\ (billion)\end{tabular}} & \textbf{\begin{tabular}[c]{@{}c@{}}Parameters\\ (million)\end{tabular}} \\ \hline
		AlexNet & 19.8 & 14.56 & 8 & 0.7 & 61 \\ \hline
		GoogleNet & 10.07 & 39.14 & 22 & 1.6 & 6.9 \\ \hline
		VGG-16 & 8.8 & 128.62 & 16 & 15.3 & 138 \\ \hline
		ResNet-50 & 7.02 & 103.58 & 50 & 3.8 & 25.6 \\ \hline
		ResNet-152 & 6.16 & 217.91 & 152 & 11.3 & 60.2 \\ \hline
	\end{tabular}
}
\vspace{2mm}
\caption{Memory and computational expensiveness of some of the most commonly used DNN models.}
	\label{tab.CNN_examples}
\end{table}

To address this challenge, the opportunities lie at exploiting the \textit{redundancy} of DNN models in terms of parameter representation and network architecture.
%
%
In terms of parameter representation redundancy, to achieve the highest accuracy, state-of-the-art DNN models routinely use 32 or 64 bits to represent model parameters.
However, for many tasks like object classification and speech recognition, such high-precision representations are not necessary and thus exhibit considerable redundancy.
Such redundancy can be effectively reduced by applying parameter quantization techniques which use 16 bits, 8 bits, or even less number of bits to represent model parameters.
%
%
In terms of network architecture redundancy, state-of-the-art DNN models use overparameterized network architectures and thus many of their parameters are redundant. 
To reduce such redundancy, the most effective technique is model compression.
In general, DNN model compression technique can be grouped into two categories. 
The first category focuses on compressing large DNN models that are pretrained into smaller ones.
For example, \cite{han2015deep} proposed a model compression technique that prunes out unimportant model parameters whose values are lower than a threshold. 
However, although this parameter pruning approach is effective at reducing model sizes, it does not necessarily reduce the number of operations involved in the DNN model.
To overcome this issue, \cite{li2016pruning} proposed a model compression technique that prunes out unimportant filters which effectively reduces the computational cost of DNN models.
%
The second category focuses on designing efficient small DNN models directly. 
For example, \cite{howard2017mobilenets} proposed to use depth-wise separable convolutions that are small and computation-efficient to replace conventional convolutions that are large and computational expensive, which reduces not only model size but also computational cost.
Being as an orthogonal approach, \cite{hinton2015distilling} proposed a technique referred to as knowledge distillation to directly extract useful knowledge from large DNN models and pass it to a smaller model which achieves similar prediction performance as the large models but with much less model parameters and computational cost.



\subsection{\textbf{Data Discrepancy in Real-World Settings}}

\begin{figure}[t]
	\centering
	\includegraphics[scale=0.55]{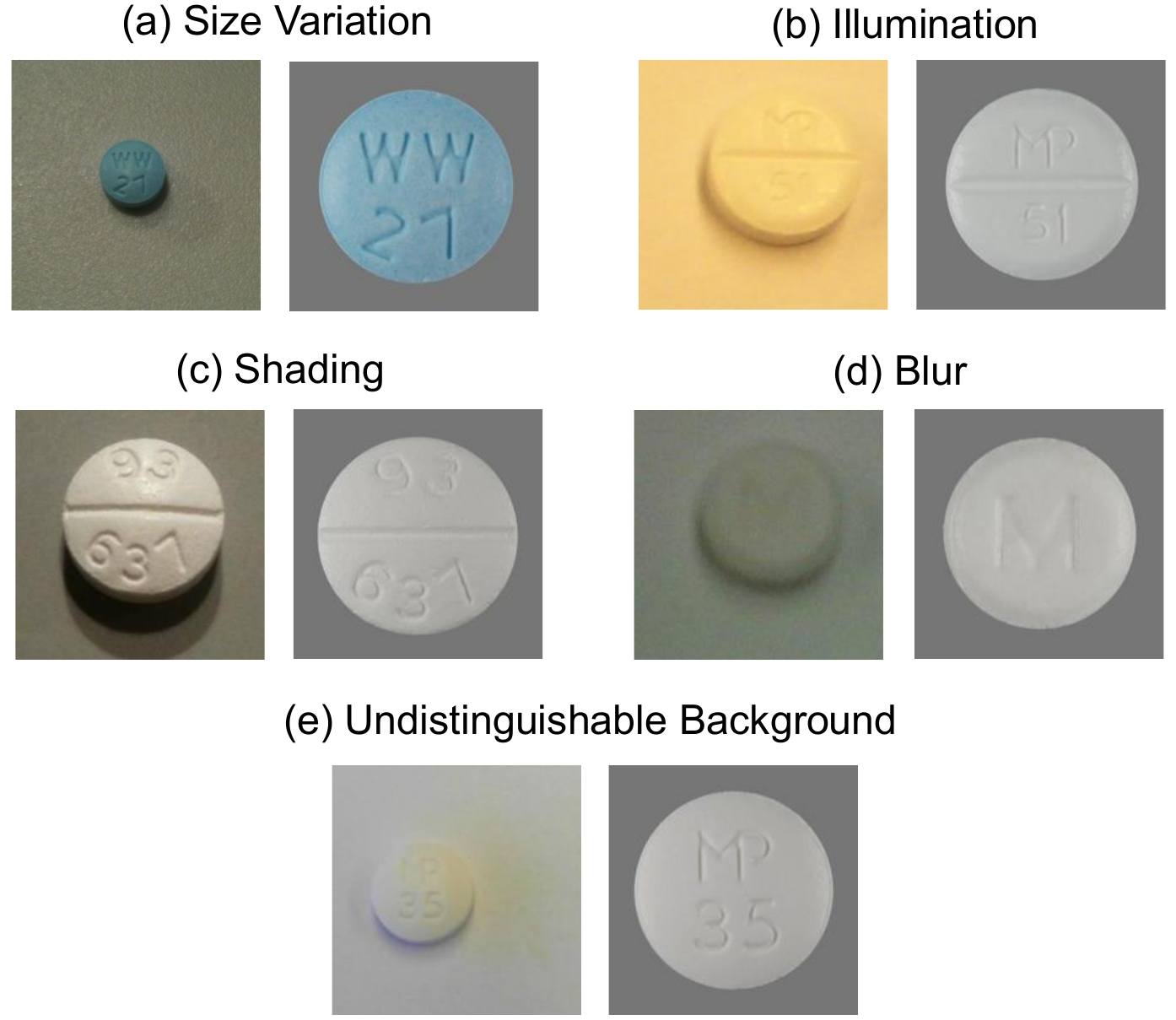}
	\caption{Illustration of differences between training and test images of the same pills under five different scenarios~\cite{zeng2017mobiledeeppill}. For each scenario, the image on the left is the training image; and the image on the right is the test image of the same pill. Due to the deterioration caused by a variety of real-world noisiness such as shading, blur, illumination and background, training image and test image of the same pill look very different.}
	\vspace{-3mm}
	\label{fig.challenges}
\end{figure}

%
The performance of a DNN model is heavily dependent on its training data, which is supposed to share the same or a similar distribution with the potential test data.
Unfortunately, in real-world settings, there can be a considerable \textit{discrepancy} between the training data and the test data.
%
Such discrepancy can be caused by variation in sensor hardware of edge devices as well as various noisy factors in the real world that degrade the quality of the test data.
%
For example, the quality of images taken in real-world settings can be degraded by factors such as illumination, shading, blurriness, and undistinguishable background~\cite{zeng2017mobiledeeppill} (see Figure \ref{fig.challenges} as an example).
Speech data sampled in noisy places such as busy restaurants can be contaminated by voices from surround people.
The discrepancy between training and test data could degrade the performance of DNN models, which becomes a challenging problem.

To address this challenge, we envision that the opportunities lie at exploring data augmentation techniques as well as designing noise-robust loss functions.
%
Specifically, to ensure the robustness of DNN models in real-world settings, a large volume of training data that contain significant variations is needed. 
Unfortunately, collecting such a large volume of diverse data that cover all types of variations and noise factors is extremely time-consuming. 
One effective technique to overcome this dilemma is data augmentation.
%
Data augmentation techniques generate variations that mimic the variations occurred in the real-world settings. 
By using the large amount of newly generated augmented data as part of the training data, the discrepancy between training and test data is minimized.
As a result, the trained DNN models become more robust to the various noisy factors in the real world.
%
%
%
A complementary technique to data augmentation is to design loss functions that are robust to discrepancy between the training data and the test data.
Examples of such noise-robust loss functions include triplet loss \cite{DBLP:journals/corr/SchroffKP15} and variational autoencoder \cite{kingma2013auto}.
These noise-robust loss functions are able to enforce a DNN model to learn features that are invariant to various noises that degrade the quality of test data even if the training data and test data do not share a similar distribution.



\noindent
\subsection{\textbf{Constrained Battery Life of Edge Devices}}

For edge devices that are powered by batteries, reducing energy consumption is critical to extending devices' battery lives.
However, some sensors that edge devices heavily count on to collect data from individuals and the physical world such as cameras are designed to capture high-quality data, which are power hungry.
%
For example, video cameras incorporated in smartphones today have increasingly high resolutions to meet people's photographic demands. 
%
As such, the quality of images taken by smartphone cameras is comparable to images that are taken by professional cameras, and image sensors inside smartphones are consuming more energy than ever before, making energy consumption reduction a significant challenge.

To address this challenge, we envision that the opportunities lie at exploring smart data subsampling techniques, matching data resolution to DNN models, and redesigning sensor hardware to make it low-power. 
First, to reduce energy consumption, one commonly used approach is to turn on the sensors when needed.
However, there are streaming applications that require sensors to be always on.
As such, it requires DNN models to be run over the streaming data in a continuous manner.
%
To reduce energy consumption in such scenario, opportunities lie at subsampling the streaming data and only processing those informative subsampled data points while discarding data points that contain redundant information.

Second, while sensor data such as raw images are high-resolution, DNN models are designed to process images at a much lower resolution (e.g., $224\times224$).
%
The mismatch between high-resolution raw images and low-resolution DNN models incurs considerable unnecessary energy consumption, including energy consumed to capture high-resolution raw images and energy consumed to convert high-resolution raw images to low-resolution ones to fit the DNN models.
To address the mismatch, one opportunity is to adopt a dual-mode mechanism.
The first mode is traditional sensing mode for photographic purposes that captures high-resolution images.
The second mode is DNN processing mode that is optimized for deep learning tasks. 
Under this model, the resolutions of collected images are enforced to match the input requirement of DNN models.
%

%
Lastly, to further reduce energy consumption, another opportunity lies at redesigning sensor hardware to reduce the energy consumption related to sensing.
When collecting data from onboard sensors, a large portion of the energy is consumed by the analog-to-digital converter (ADC).
%
There are early works that explored the feasibility of removing ADC and directly using analog sensor signals as inputs for DNN models \cite{likamwa2016redeye}. Their promising results demonstrate the significant potential of this research direction.




\subsection{\textbf{Heterogeneity in Sensor Data}}
Many edge devices are equipped with more than one onboard sensor.
For example, a smartphone has a GPS sensor to track geographical locations, an accelerometer to capture physical movements, a light sensor to measure ambient light levels, a touchscreen sensor to monitor users’ interactions with their phones, a microphone to collect audio information, and a camera to capture images and videos.
Data obtained by these sensors are by nature \textit{heterogeneous} and are diverse in format, dimensions, sampling rates, and scales.
%
How to take the data heterogeneity into consideration to build DNN models and to effectively integrate the heterogeneous sensor data as inputs for DNN models represent a significant challenge.

To address this challenge, one opportunity lies at building a multi-modal deep learning model that takes data from different sensing modalities as its inputs.
For example, \cite{radu2016towards} proposed a multi-modal DNN model that uses Restricted Boltzmann Machine for activity recognition. 
Similarly, \cite{bhattacharya2016smart} also proposed a multi-modal DNN model for smartwatch-based activity recognition. 
Besides building multi-modal DNN models, another opportunity lies at combining information from heterogeneous sensor data extracted at different dimensions and scales.
As an example, \cite{chang2015heterogeneous} proposed a multi-resolution deep embedding approach for processing heterogeneous data at different dimensions.
\cite{DBLP:journals/corr/YaoHZZA16} proposed an integrated convolutional and recurrent neural networks for processing heterogeneous data at different scales. 
%


\subsection{\textbf{Heterogeneity in Computing Units}}
\vspace{-1mm}
Besides data heterogeneity, edge devices are also confronted with heterogeneity in on-device computing units.
As computing hardware becomes more and more specialized, an edge device could have a diverse set of onboard computing units including traditional processors such as CPUs, DSPs, GPUs, and FPGAs as well as emerging domain-specific processors such as Google's Tensor Processing Unit (TPUs).
Given the increasing heterogeneity in onboard computing units, mapping deep learning tasks and DNN models to the diverse set of onboard computing units is challenging.

To address this challenge, the opportunity lies at mapping operations involved in DNN model executions to the computing unit that is optimized for them.
State-of-the-art DNN models incorporate a diverse set of operations but can be generally grouped into two categories: parallel operations and sequential operations.
For example, the convolution operations involved in convolutional neural networks (CNNs) are matrix multiplications that can be efficiently executed in parallel on GPUs which have the optimized architecture for executing parallel operations.
In contrast, the operations involved in recurrent neural networks (RNNs) have strong sequential dependencies, and better fit CPUs which are optimized for executing sequential operations where operator dependencies exist.
%
The diversity of operations suggests the importance of building an architecture-aware compiler that is able to decompose a DNN models at the operation level and then allocate the right type of computing unit to execute the operations that fit its architecture characteristics.
%
Such an architecture-aware compiler would maximize the hardware resource utilization and significantly improve the DNN model execution efficiency. 

%



\subsection{\textbf{Multi-Tenancy of Deep Learning Tasks}}
\vspace{-1mm}
%
The complexity of real-world applications requires edge devices to concurrently execute multiple DNN models which target different deep learning tasks~\cite{fangzeng2018nestdnn}.
%
For example, a service robot that needs to interact with customers needs to not only track faces of individuals it interacts with but also recognize their facial emotions at the same time. 
These tasks all share the same data inputs and the limited resources on the edge device.
How to effectively share the data inputs across concurrent deep learning tasks and efficiently utilize the shared resources to maximize the overall performance of all the concurrent deep learning tasks is challenging.

\begin{figure}[t]
	\begin{center}
		\includegraphics[scale=0.6,clip=true,trim = 85mm 63mm 100mm 50mm]{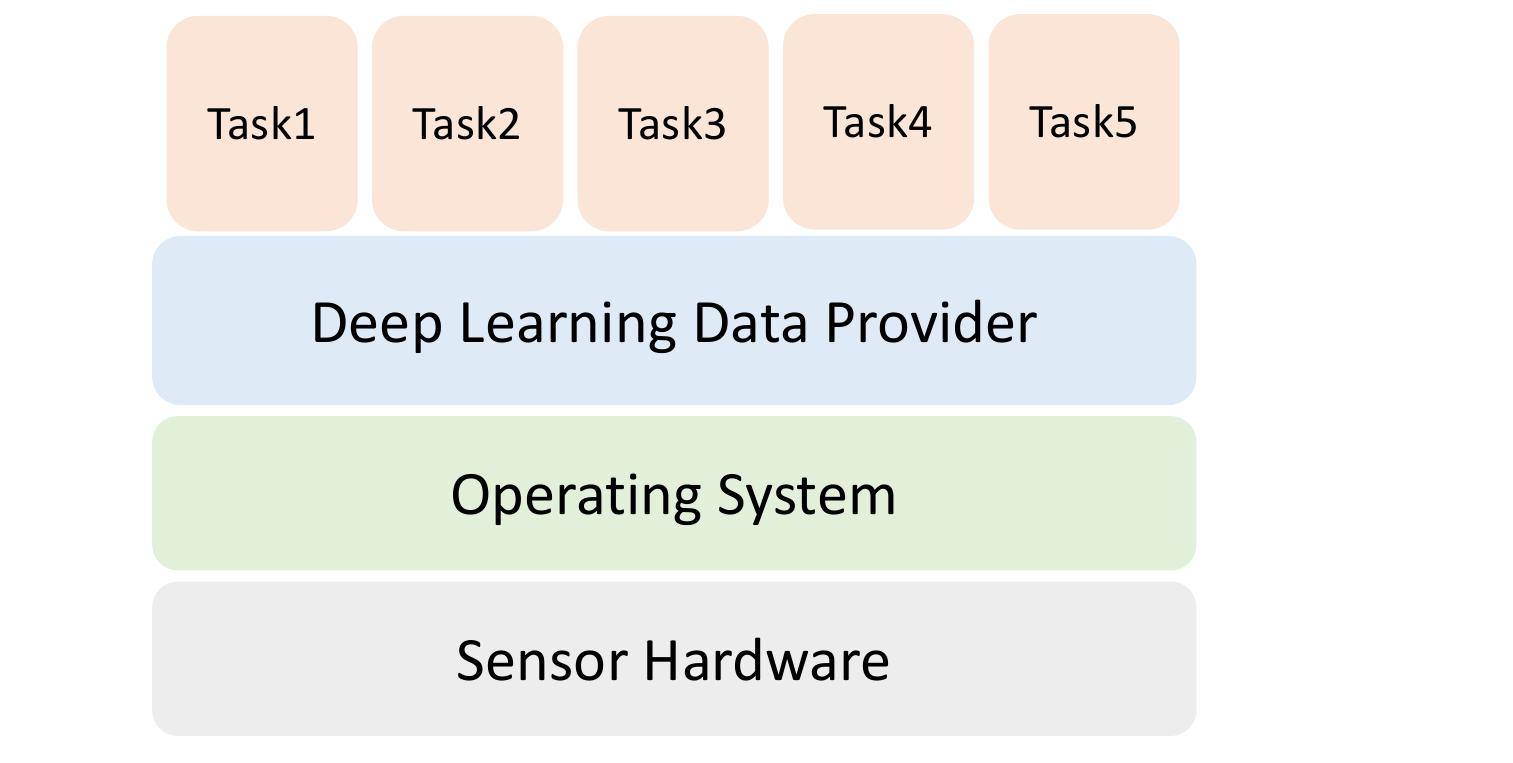}
	\end{center}
	\vspace{-4mm}
	\caption{Illustration of data sharing mechanism.} 
	\label{fig.data_sharing}
	\vspace{-4mm}
\end{figure}

In terms of input data sharing, currently, data acquisition for concurrently running deep learning tasks on edge devices is exclusive. 
In other words, at runtime, only one single deep learning task is able to access the sensor data inputs at one time.
%
As a consequence, when there are multiple deep learning tasks running concurrently on edge devices, each deep learning task has to explicitly invoke system APIs to obtain its own data copy and maintain it in its own process space.
This mechanism causes considerable system overhead as the number of concurrently running deep learning tasks increases.
To address this input data sharing challenge, one opportunity lies at creating a \textit{data provider} that is transparent to deep learning tasks and sits between them and the operating system as shown in Figure \ref{fig.data_sharing}.
The \textit{data provider} creates a single copy of the sensor data inputs such that deep learning tasks that need to acquire data all access to this single copy for data acquisition.
As such, a deep learning task is able to acquire data without interfering other tasks. 
More importantly, it provides a solution that scales in terms of the number of concurrently running deep learning tasks.

%

%
In terms of resource sharing, in common practice, DNN models are designed for individual deep learning tasks. 
However, existing works in deep learning show that DNN models exhibit layer-wise semantics where bottom layers extract basic structures and low-level features while layers at upper levels extract complex structures and high-level features. 
%
This key finding aligns with a sub-field in machine learning named multi-task learning~\cite{caruana1997multitask}. 
In multi-task learning, a single model is trained to perform multiple tasks by sharing low-level features while high-level features differ for different tasks.
For example, a DNN model can be trained for scene understanding as well as object classification~\cite{zhou2014object}.
Multi-task learning provides a perfect opportunity for improving the resource utilization for resource-limited edge devices when concurrently executing multiple deep learning tasks.
By sharing the low-level layers of the DNN model across different deep learning tasks, redundancy across deep learning tasks can be maximally reduced.
%
In doing so, edge devices can efficiently utilize the shared resources to maximize the overall performance of all the concurrent deep learning tasks.
%




\subsection{\textbf{Offloading to Nearby Edges}}
\vspace{-2mm}
For edge devices that have extremely limited resources such as low-end IoT devices, they may still not be able to afford executing the most memory and computation-efficient DNN models locally.
In such scenario, instead of running the DNN models locally, it is necessary to offload the execution of DNN models.
As mentioned in the introduction section, offloading to the cloud has a number of drawbacks, including leaking user privacy and suffering from unpredictable end-to-end network latency that could affect user experience, especially when real-time feedback is needed.
Considering those drawbacks, a better option is to offload to nearby edge devices that have ample resources to execute the DNN models.

\begin{figure}[!ht]
	\begin{center}
		\includegraphics[scale=0.5,clip=true,trim = 100mm 60mm 100mm 50mm]{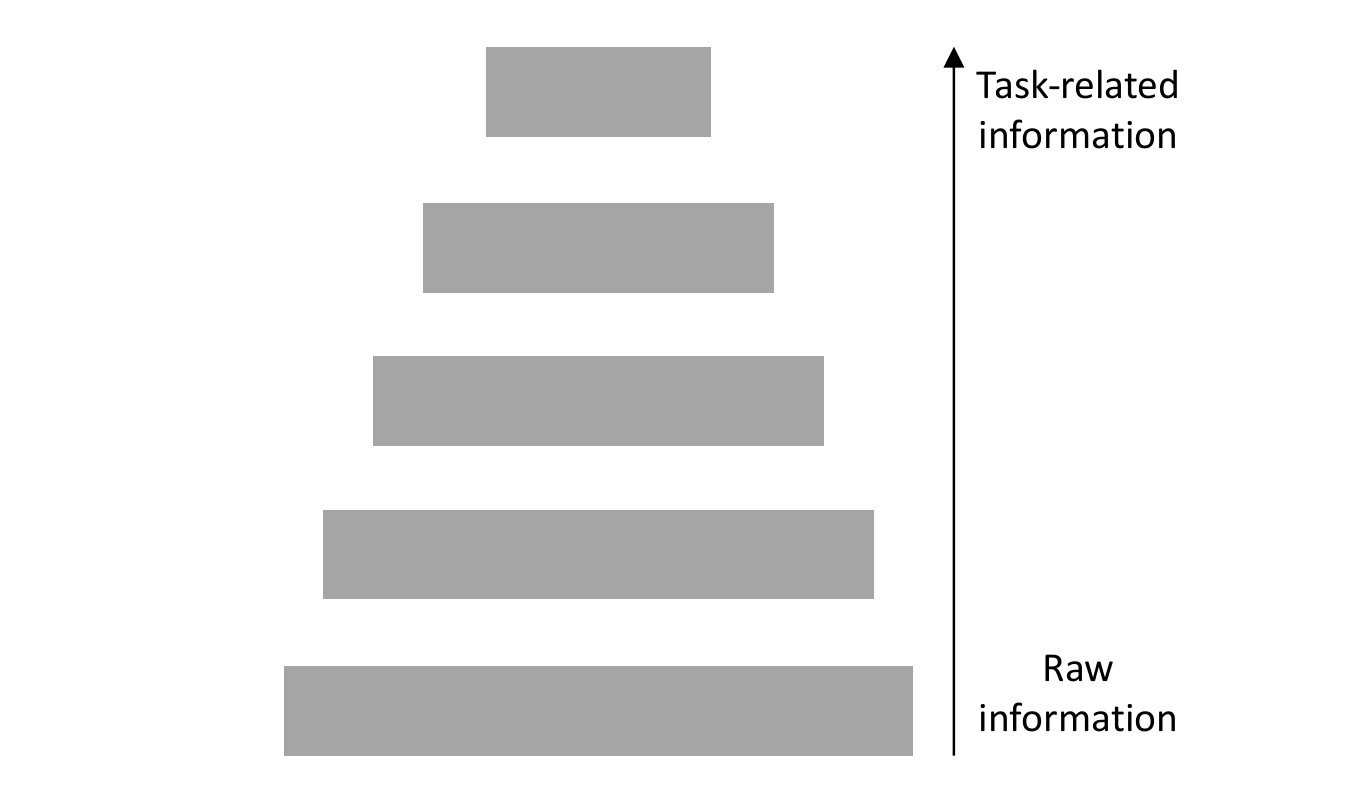}
	\end{center}
	\vspace{-4mm}
	\caption{Illustration of intermediate results of a DNN model. The size of intermediate results generated out of each layer decreases from lower layers to higher layers. The amount of information generated out of each layer also decreases from lower layers to higher layers. } 
	\label{fig.pyramid}
	\vspace{-4mm}
\end{figure}

To realize edge offloading, the key is to come up with a model partition and allocation scheme that determines which part of model should be executed locally and which part of model should be offloading.
%
To answer this question, the first aspect that needs to take into account is the size of intermediate results of executing a DNN model.
%
A DNN model adopts a layered architecture. 
The sizes of intermediate results generated out of each layer have a pyramid shape (Figure \ref{fig.pyramid}), decreasing from lower layers to higher layers. 
%
As a result, partitioning at lower layers would generate larger sizes of intermediate results, which could increase the transmission latency.
%
The second aspect that needs to take into account is the amount of information to be transmitted.
%
For a DNN model, the amount of information generated out of each layer decreases from lower layers to higher layers. 
Partitioning at lower layers would prevent more information from being transmitted, thus preserving more privacy.
%
As such, the edge offloading scheme creates a trade-off between computation workload, transmission latency, and privacy preservation.

\subsection{\textbf{On-Device Training}}
\vspace{-2mm}
%
%
In common practice, DNN models are trained on high-end workstations equipped with powerful GPUs where training data are also located.
This is the approach that giant AI companies such as Google, Facebook, and Amazon have adopted. 
These companies have been collecting a gigantic amount of data from users and use those data to train their DNN models. 
This approach, however, is privacy-intrusive, especially for mobile phone users because mobile phones may contain the users’ privacy-sensitive data. 
How to best protect users' privacy while still obtaining well-trained DNN models becomes a challenging problem.

To address this challenge, we envision that the opportunity lies at on-device training.
As compute resources in edge devices become increasingly powerful, especially with the emergence of AI chipsets, in the near future, it becomes feasible to train a DNN model locally on edge devices.
%
By keeping all the personal data that may contain private information on edge devices, on-device training provides a privacy-preserving mechanism that  leverages the compute resources inside edge devices to train DNN models without sending the privacy-sensitive personal data to the giant AI companies. 
Moreover, today, gigantic amounts of data are generated by edge devices such as mobile phones on a daily basis. 
These data contain valuable information about users and their personal preferences. 
With such personal information, on-device training is enabling training personalized DNN models that deliver personalized services to maximally enhance user experiences.

%

\vspace{-2mm}
\section{Concluding Remarks}
\label{sec.con}
\vspace{-2mm}

\noindent
Edge computing is revolutionizing the way we live, work, and interact with the world.
With the recent breakthrough in deep learning, it is expected that in the foreseeable future, majority of the edge devices will be equipped with machine intelligence powered by deep learning.
To realize the full promise of deep learning in the era of edge computing, there are daunting challenges to address.

In this book chapter, we presented eight challenges at the intersection of computer systems, networking, and machine learning.
These challenges are driven by the gap between high computational demand of DNN models and the limited battery lives of edge devices, the data discrepancy in real-world settings, the needs to process heterogeneous sensor data and concurrent deep learning tasks on heterogeneous computing units, and the opportunities for offloading to nearby edges and on-device training.
We also proposed opportunities that have potential to address these challenges.
%
We hope our discussion could inspire new research that turns the envisioned intelligent edge into reality.

\bibliographystyle{plain}
\bibliography{sigproc}

\end{document}